\documentclass[sageapa,times,Crown]{sagej}

\usepackage{moreverb,url}
\usepackage[colorlinks,bookmarksopen,bookmarksnumbered,citecolor=red,urlcolor=red]{hyperref}

\usepackage{xcolor}
\usepackage{multirow}
\usepackage{booktabs}

\usepackage{amssymb}
\usepackage{caption}
\usepackage{subcaption}
\usepackage{graphicx}
\usepackage{amsmath}
\usepackage{float}
\usepackage[normalem]{ulem}
\usepackage{enumitem}
\usepackage{adjustbox}
\usepackage{lineno}

\newcommand\BibTeX{{\rmfamily B\kern-.05em \textsc{i\kern-.025em b}\kern-.08em
T\kern-.1667em\lower.7ex\hbox{E}\kern-.125emX}}

\def\volumeyear{2016}

\definecolor{green}{rgb}{0.0, 0.65, 0.31}

\definecolor{bleudefrance}{rgb}{0.19, 0.55, 0.91}

\newcommand{\multrow}[1]{\begin{tabular}{@{}c@{}} #1 \end{tabular}}

\newboolean{blinded}
\setboolean{blinded}{false} % true: anonymized | false: identified
\newboolean{coverpage}
\setboolean{coverpage}{false} % true: only coverage | false: full paper

\ifthenelse{\boolean{blinded}}
{
    \linenumbers
}{
}

\ifthenelse{\boolean{coverpage}}
{
\setboolean{blinded}{false} % true: anonymized | false: identified
}{
}

\begin{document}

\title{Explainable Artificial Intelligence for Quantifying Interfering and High-Risk Behaviors in Autism Spectrum Disorder in a Real-World Classroom Environment Using Privacy-Preserving Video Analysis}

\ifthenelse{\boolean{blinded}}
{
}{
    \runninghead{Das \textit{et al.}}
    \author{Barun Das\affilnum{1}, 
    Conor Anderson\affilnum{2}, 
    Tania Villavicencio\affilnum{2}, 
    Johanna Lantz\affilnum{2}, 
    Jenny Foster\affilnum{2}, 
    Theresa Hamlin\affilnum{2}, 
    Ali Bahrami Rad\affilnum{1}, 
    Gari D. Clifford\affilnum{1,3,*}, and 
    Hyeokhyen Kwon\affilnum{1,3,*}}
    \affiliation{\affilnum{1}Department of Biomedical Informatics, Emory University School of Medicine, Atlanta, GA\\ 
    \affilnum{2}The Center for Discovery, New York, NY\\
    \affilnum{3}Department of Biomedical Engineering, Emory University and Georgia Institute of Technology, Atlanta, GA\\
    \affilnum{*}Joint Senior Authors
    }
    \corrauth{Hyeokhyen Kwon, Department of Biomedical Informatics, Emory University School of Medicine, Atlanta, GA.}
    \email{hyeokhyen.kwon@emory.edu}
}

\ifthenelse{\boolean{coverpage}}
{
    \maketitle
}{
    \begin{abstract}
    \textit{
    Rapid identification and accurate documentation of interfering and high-risk behaviors in ASD, such as aggression, self-injury,
disruption, and restricted repetitive behaviors, are important in daily classroom environments for tracking intervention effectiveness and allocating appropriate resources to manage care needs.
However, having a staff dedicated solely to observing is costly and uncommon in most educational settings.
      Recently, multiple research studies have explored developing automated, continuous, and objective tools using machine learning models to quantify behaviors in ASD.
      However, the majority of the work was conducted under a controlled environment and has not been validated for real-world conditions.
      In this work, we demonstrate that the latest advances in video-based group activity recognition techniques can quantify behaviors in ASD in real-world activities in classroom environments while preserving privacy.
      Our explainable model could detect the episode of problem behaviors with a 77\% F1-score and capture distinctive behavior features in different types of behaviors in ASD.
      % Our attention method could also identify the specific child exhibiting the problem behavior within the groups.
      To the best of our knowledge, this is the first work that shows the promise of objectively quantifying behaviors in ASD in a real-world environment, which is an important step toward the development of a practical tool that can ease the burden of data collection for classroom staff. 
}

    \end{abstract}
    
    
    \maketitle
    
    % \hyeok{Overall comment across all pages. your paragraphs are too short.  Avoid having short paragraphs. Try to have a rectangular paragraph where two equal-sized squares can fit in or a little longer than that. Also, citations are missing A LOT.}

\section*{Introduction}

Autism spectrum disorder (ASD) is a neurodevelopmental disorder that is defined by difficulties in social communication and interaction due to the presence of restricted, repetitive patterns of behavior, interests, or activities~\cite{edition2013diagnostic}.   Despite these core characteristics, there is significant heterogeneity in presentation. Some autistic people lead independent and fulfilling lives, while others are more profoundly affected. Those with profound autism require 24-hour supervision, have limited language and intellectual impairment, and commonly have co-occurring medical or psychiatric conditions. While the cause of ASD is unknown and no diagnostic biomarker exists, there are several genetic and environmental risk factors associated with the condition~\cite{sauer2021autism}.

Many children with ASD present with behaviors such as aggression, self-injurious behaviors (SIB), and restrictive, repetitive behaviors (RRB) that interfere with daily functioning and may be potentially dangerous.  The standard practice for assessing interfering behaviors is to conduct a functional behavior assessment (FBA) to determine contributing factors, antecedents, and consequences. The first step of this process is to analyze the properties of the behavior such as topography, frequency, duration and intensity. Once these are characterized, interventions are developed to reduce or eliminate behaviors that interfere with functioning or could produce harm to self or others. 

Accurate data collection is critical for behavior assessment, monitoring behavior status, and analyzing the effectiveness of interventions; however, observations by behavior clinicians or classroom staff can also be subjective and inconsistent \cite{taylor2017} as certain autism behaviors have lower inter-rater reliability. \cite{matson2008asdirr}. Real-time data collection in classroom settings can be burdensome for staff who must manage other responsibilities while observing behaviors, reducing the accuracy of data reporting. Furthermore, it may not always be possible for staff to step away from supervision to record data in a timely manner, which can result in recall bias. Automatic detection of behavioral events has the potential to be more efficient and less prone to error. 

% , being able to predict such behavioral episodes is  helpful to parents, teachers and guardians in providing specific interventions.
% These are especially useful for social environments such as classrooms where these behaviors can be disruptive.

% Studying behaviors in autism is an arduous process involving numerous visits to a developmental disorder specialist and a series of observations and screening tests.

With advances in passive sensing technologies, such as wearables, high resolution cameras, nose canceling microphonees, and artificial intelligence, multiple works have explored developing computational behavior analysis systems that can automatically detect behavior or screen for diagnostic features of ASD in children.
These include using wearable sensors for quantifying repetitive behaviors,~\cite{plotz2012automatic,gilchrist2018automated} or  identifying differences in speech prosody commonly seen in children with ASD.~\cite{wijesinghe2019machine,lau2022cross,chi2022classifying}.
Other works explored using videos for capturing behaviors in those with ASD, including quantifying self-stimulatory behaviors, such as hand clapping or stimming~\cite{math11194208,de2020computer,rgopalan2013stim}.
Facial expression and head movements have also analyzed to screen for an ASD diagnosis while children watched developmentally appropriate visual stimuli or participated in response-to-name tests~\cite{hashemi2015,bidwell2014measuring,wang2021rtn}.
Whole body movements extracted from videos have been found to be useful in detecting behaviors associated with ASD~\cite{vyas2019recognition}.
In particular, a few works have explored quantifying social interaction using computer vision and audio analysis~\cite{rehg2013decoding,rgopalan2013stim}. 

Although existing research looks promising, most studies have utilized techniques that were validated in a controlled lab environment, focusing on a single individual or dyadic interaction analysis. This limits the generalizability of results and translation to practice. In this work, we evaluated whether ambient cameras and state-of-the-art computer vision models could objectively and automatically quantify behaviors while students with ASD take part in group activities in a real-world classroom environment.
To the best of our knowledge, this is the first work to demonstrate that rich movement and activity information available from ambient cameras can be used to automatically, objectively, and passively quantify target behaviors in real-world classroom environments, where children with ASD, staff, and teachers have complex social interactions.
The outcome of this work is an important step toward longitudinal and large-scale monitoring of behaviors in children with ASD.

    \section*{Method \& Materials}
% \hyeok{Changed the title to Method, following the template.}

\subsection{Setting}
% \textbf{\textit{Dataset}}

% \textcolor{orange}{
%     Questions to CFD team:
%     \begin{itemize}
%         \item Study site details (area, location, etc.)
%         \item How long do the subjects stay at school?
%         \item Total number of subjects, subjects' age, sex, race (demographic information)
%         \item Do the children only interact among themselves and the teaching staff?
%         % \item Data collection procedures (data transfer?) \hyeok{This will be just a camera recording from two views, which we actually report in this paper.}
%         \item Annotation details. How many annotators? Is there any interrater reliability analysis with multiple annotators? I believe the annotation was done retrospectively. Is this correct?
%         % \hyeok{You already have this from the annotation readme shared by CFD} \barun{Not completely. Need more details. How many annotators? Was it annotated real-time or retrospectively? Were annotations cross-checked amongst experts?}
%     \end{itemize}
% }

% \input{tables/frames-labeled}

% \sout{The study site is fitted with a camera in top-down view, as shown in \ref{fig:camera-views}, that captures a video feed of the classroom, including the students and staff.}
% \hyeok{This contradicts your main result, Table 1, which reports both Cam 1 and 2. Since you want to show both results and discuss them. Mention that you are using two cameras from a top-down and side view. You also have to move your Figure 3 to here and refer it, which I did, since this is the first paragraph going into detail about how your data is collected.}

We captured classroom activities of students attending The Center for Discovery (TCFD), Harris, New York, a  research and specialty center that offers residential, medical, clinical and special education programs for medically and behaviorally complex individuals, including those with profound ASD and other intellectual and developmental disabilities. Participants in this study included both day and residential students in the education program. The classroom ratio included six children, one teacher, and three teacher associates. Additional support staff such as behavior analysts, speech and language pathologists, and occupational therapists were frequently present in the classroom. Although there were six children enrolled at a given time, classes were often broken into groups so that a particular lesson only included three to four students. The classroom utilized for this study was designated as a research classroom with enhanced data collection procedures and was instrumented with cameras. Despite these additions, the classroom functioned as a typical special education class with lessons and instructional activities. 

\subsection{Participants}

\begin{table}[t]
    \centering
    \caption{Scores and classifications from the Autism Spectrum Rating Scale (ASRS) and adaptive behavior scales including the Vineland Adaptive Behavior Scales- Second Edition (VABS 2) and the Adaptive Behavior Assessment System- 2nd Edition (ABAS 2).}
    \begin{tabular}{c||ccc}
         Subject ID & \multrow{ASRS DSM Score\\(Classification)} & \multrow{Adaptive Behavior Score\\(Classification)} & Adaptive behavior Measure \\
         \hline\hline
         S01 & 83 (Very Elevated) & 40 (Low) & VABS 2\\
         \hline
         S03 & 80 (Very Elevated) & 37 (Low) & VABS 2\\
         \hline
         S06 & 84 (Very Elevated) & 45 (Low) & VABS 2\\
         \hline
         S07 & 80 (Very Elevated) & 40 (Low) & VABS 2\\
         \hline
         S08 & 85 (Very Elevated) & 38 (Low) & VABS 2\\
         \hline
         S09 & 75 (Very Elevated) & 48 (Low) & VABS 2\\
         \hline
         S10 & 57 (Average) & 23 (Low) & VABS 2\\
         \hline
         S11 & 77 (Very Elevated) & 35 (Low) & VABS 2\\
         \hline
         S12 & 70 (Very Elevated) & 48 (Extremely Low) & ABAS 2\\
         \hline
         S13 & 73 (very Elevated) & 41 (Extremely Low)  & ABAS 2
    \end{tabular}
    \label{tab:demograph}
\end{table}

Data were analyzed for nine students ages 12 to 20 who participated in the research classroom during a four-year period. All students were males previously diagnosed with ASD and Moderate to Severe Intellectual Disability. As part of standard procedures, the severity of ASD symptoms is assessed annually until the age of 18 using the Autism Spectrum Rating Scale (ASRS) \cite{goldstein2009autism}. For participants who were over the age of 18 during the study, scores from the last ASRS administered were included.  As can be seen in \autoref{tab:demograph}, all students displayed "Very Elevated" symptoms of autism except for one (S10), whose presentation was "Average" compared to other youths with ASD. Due to limited communication and difficulties performing novel tasks that impeded the ability to participate in cognitive testing,  adaptive behavior measures were administered to assess the communication, socialization, and daily living skills of participants in this study. Adaptive behavior measures included the Vineland Adaptive Behavior Scales- 2nd Edition (VABS-2)~\cite{sparrow2005vineland} or the Adaptive Behavior Assessment System- 2nd edition (ABAS-2)~\cite{harrison2000adaptive}. All scores fell in the lowest classification range, indicating severe deficits in all areas assessed.  

\subsection{Data Collection Procedures}
All students in the TCFD research classroom had behaviors targeted for reduction that were operationally defined by behavior analysts. These "target" behaviors were those that interfered with an individual's learning, that were disruptive to others, or that were considered high-risk due to the potential to cause injury or property destruction. Restricted and repetitive behaviors such as loud vocalizations or repetitive body movements were targeted for intervention only if they were problematic to other students due to the noise level or visual distraction. 

The TCFD research classroom  was instrumented 6 cameras, all with different angles: Camera 1 top view wide angle, Camera 2 side view wide angle and Cameras 3-6 with side angles. 
Video and audio recording was captured using the multi-camera video recording software StreamPix  (Norpix, Montreal, QC, Canada).
% \footnote{https://www.norpix.com/products/streampix/streampix.php}. 
For the current study, Cameras 1 and 2 were selected, as both views capture the students and staff interactions with greater detail, as shown in \autoref{fig:camera-views}. These cameras capture the interactions between the teaching staff and students with ASD during the class. 

Trained research assistants extracted and manually annotated episodes of the target behaviors from video recordings. The length of these video recordings averaged 25 minutes for each class session. In this study, 139 videos were analyzed (69 top-down and 70 side views). Research assistants annotated the videos using Noldus The Observer XT 15 software (Wageningen, the Netherlands), labeling each target behavior's onset and offset times. 
In total, there were 21.9 hours of labeled videos for Camera 1 (top-down view) and 22.2 hours for Camera 2 (side view), and the frame rate of the camera was 30 Hz.
The annotated episodes of target behaviors included
\textit{restricted repetitive behaviors, 
self-injurious behaviors, 
disruptive behaviors, 
aggressive behaviors, 
elopement},
and 
\textit{out-of-seat} 
during class activities.

% From there, we extract 2D pose keypoints of the individuals observed within the scene using Posenet \cite{posenet}.
% In order to preserve privacy, we do not store the raw video frames. Instead, we store the Posenet-extracted 2D human poses, with each pose consisting of  X and Y coordinates of 17 keypoints \cite{posenet} 
% \hyeok{This sent
% This process involved marking each identified behavior's onset and offset times, with latency times between behaviors ranging from five to ten seconds, depending on the operational definitions established for each participant. The annotated behaviors were then exported into CSV files for further analysis.
% (\textcolor{red}{XXX} annotators with interrater reliability of \textcolor{red}{XXX}), which included N=\textcolor{red}{XXX} \barun{N=11, I believe. CFD team: please confirm} subjects. 

% In this work, we aim to detect episodes of these atypical behaviors by using a continuous, passive video analysis system for group activities in the classroom. 
% \barun{\sout{See example in https://dl.acm.org/doi/pdf/10.1145/3594738.3611364}}
% \resolved
% We codified these behaviors, assigning each of them a numeric label.
\autoref{fig:problem-behaviors} illustrates the distribution of each behavioral label observed in the dataset, and \autoref{fig:subject-behaviors} shows the distribution of behavioral episodes for each participant.
The majority of behaviors belong to \textit{restricted repetitive behavior}, \textit{disruptive behavior}, and \textit{out-of-seat behaviors}, and the participants showed highly variable behavior episodes.
For example, one participant mainly showed \textit{self-injurious} behaviors (S11), whereas another participant mainly showed more \textit{restricted repetitive behaviors} (S09).
In this study, we focused on detecting the presence of target behaviors regardless of highly heterogeneous behaviors occurring in the classroom. With the annotated videos, our proposed system models group activity in the scene using an attention-based group activity recognition system that automatically identifies the most relevant subjects presenting atypical behaviors in the scene. 
% \textcolor{orange}{Why is such a distribution observed [clinical perspective]?}

\begin{figure}[t]
    \centering
    \begin{subfigure}{0.45\textwidth}
        % \centering
        \includegraphics[width=\linewidth]{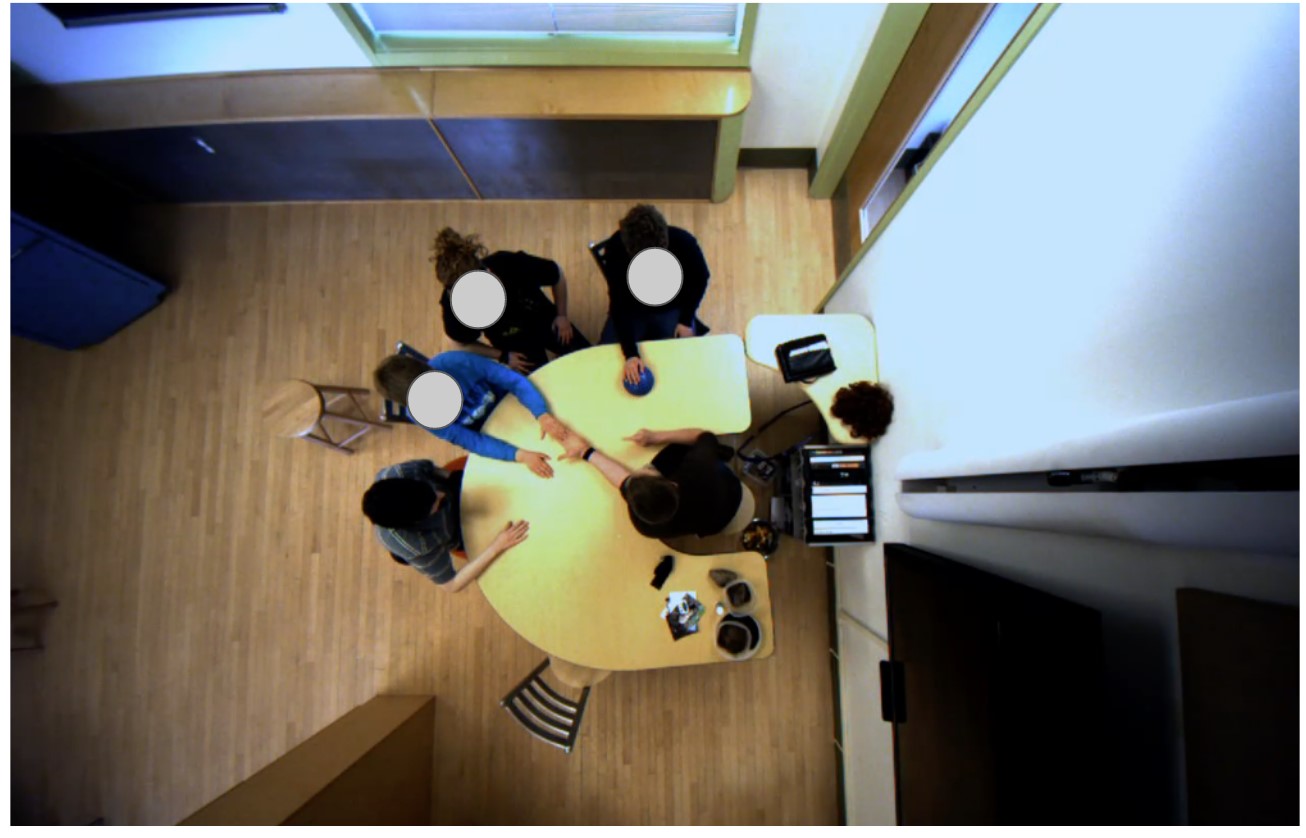}
        \caption{\small Camera 1 view \textit{(top-down)}}
        \label{fig:cam1-view}
    \end{subfigure}
    \hfill
    \begin{subfigure}{0.45\textwidth}
        % \centering
        \includegraphics[width=\linewidth]{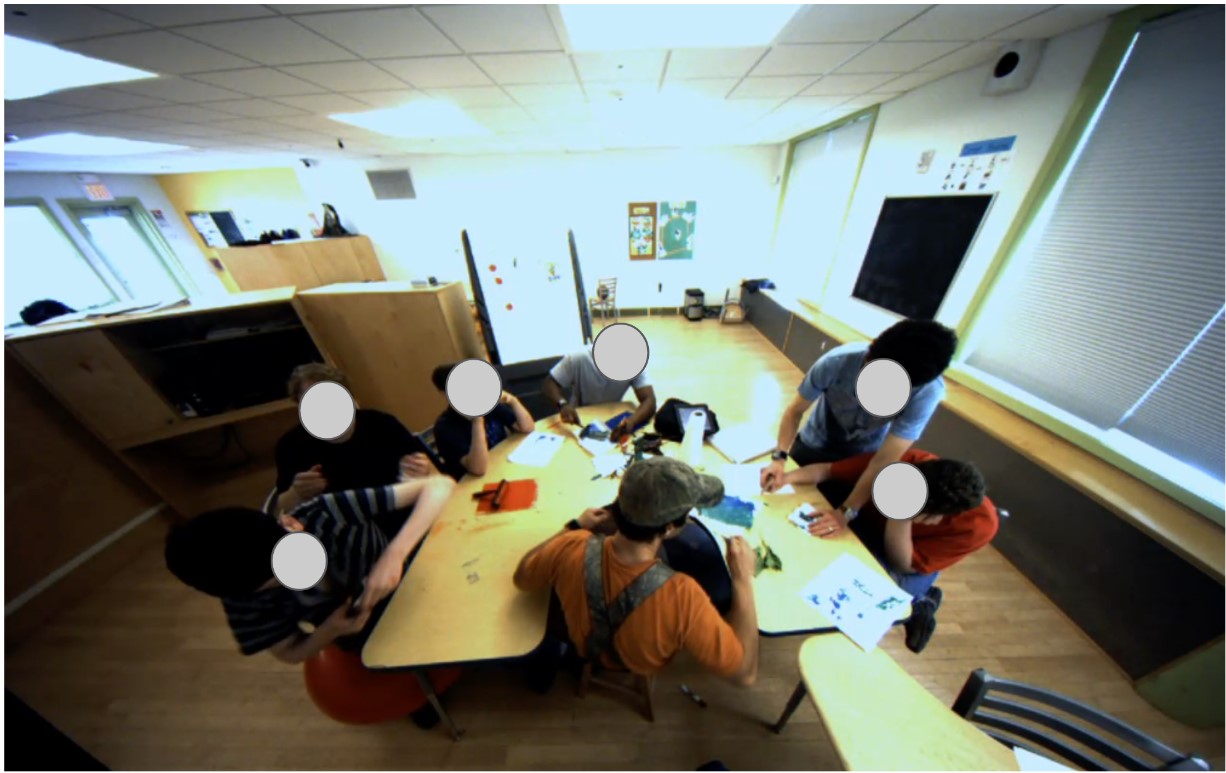}
        \caption{\small Camera 2 view \textit{(side)}}
        \label{fig:cam2-view}
    \end{subfigure}
    \hfill
    % \vspace{-1em}
    \caption{\small Two different camera views capturing the classroom table in our study. 
    % \hyeok{Needs to appear where first time discussed in detail.}
    %The participant' faces have been blotted out to protect their identities.
    }
    \label{fig:camera-views}        
\end{figure}

\begin{figure}[t!]
    \centering
    % \vspace{-2em}
    \includegraphics[width=\textwidth]{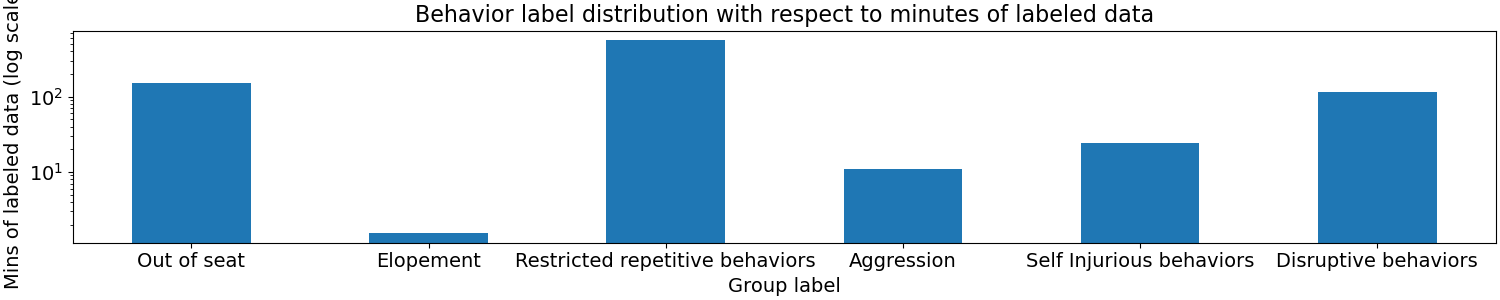}
    \caption{Distribution of total duration (in minutes) target behaviors observed in our labeled dataset. 
    }
    \label{fig:problem-behaviors}
\end{figure}

\begin{figure*}[t!]
    \centering
    \includegraphics[width=\linewidth]{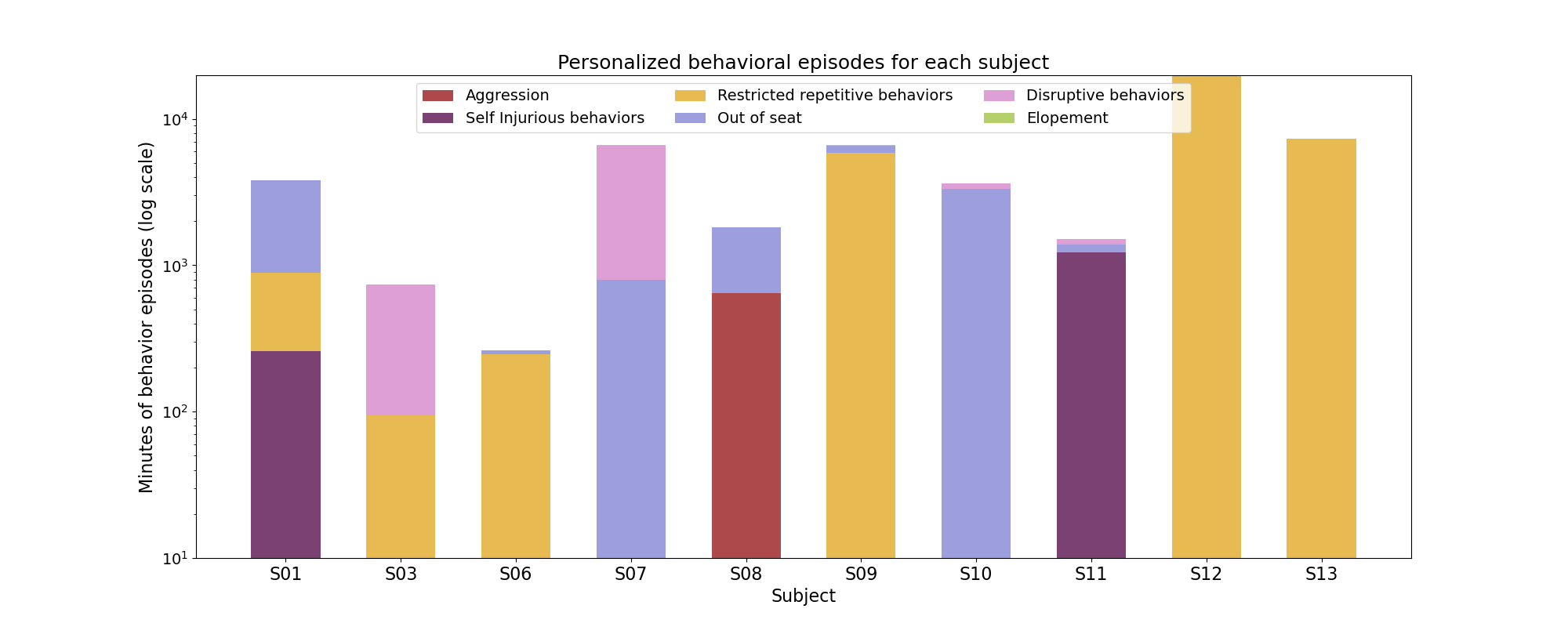}
    \vspace{-2em}
    \caption{Distribution of behavioral episodes for each subject}
    \label{fig:subject-behaviors}
\end{figure*}

\subsection{Privacy-preserving and Explainable Group Activity Analysis}
% \textit{\textbf{Privacy-preserving and Explainable Group Activity Analysis}}

\begin{figure*}[t!]
        \centering
        \includegraphics[width=1.\textwidth]{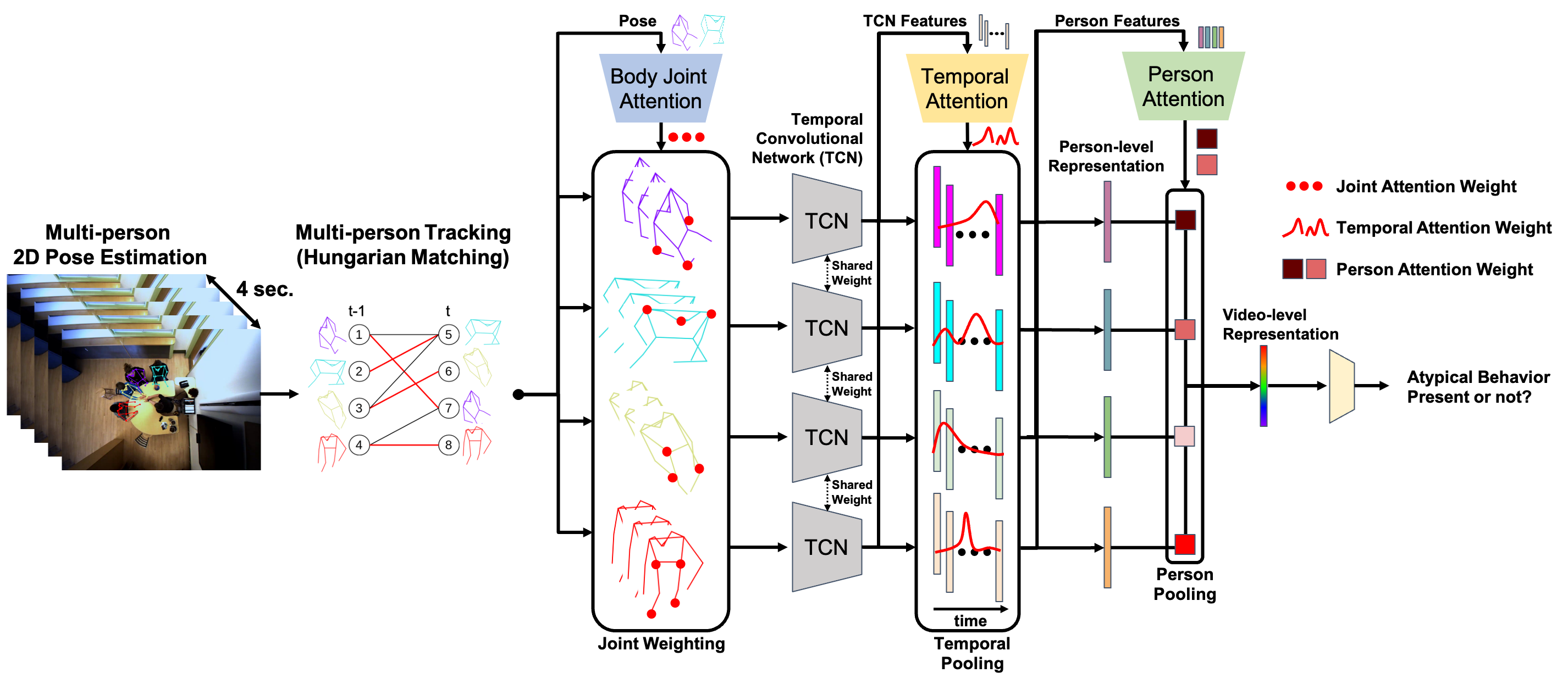}
        \caption{Overall analysis pipeline for group activities in the classroom (Example with a top-down view).
        From the 4-second analysis window, 2D poses of subjects in the scene are captured using DEKR model~\protect\cite{geng2021bottom}, and each subject's activities over the video clip are collected through multi-person tracking technique using Hungarian matching algorithm~\protect\cite{kuhn1955hungarian}. 
        The proposed method first identifies the important joint movements by processing the pose sequence using the \textit{Body Joint Attention} model.
        The joint-weighted pose sequences are then processed with a Temporal Convolutional Network (TCN) to extract pose features across all frames.
        The TCN feature sequences are then processed with \textit{Temporal Attention} model for temporal pooling with higher weights on the more relevant frames to identify atypical behaviors in the 4-second window.
        Temporally pooled features for each individual are then processed with \textit{Person Attention} model to identify the subject that is most likely to exhibit atypical behaviors.
        The person-attention weights are used to aggregate all individuals' features to generate video-level features, which are used to identify the presence of target behavior in 4-second analysis windows.   
        }  
        \label{fig:models}
\end{figure*}

\subsubsection{Overall Pipeline}

\autoref{fig:models} shows the overall pipeline of our group activity analysis system to detect episodes of target behaviors in the classroom.
We apply the standard human activity recognition pipeline~\cite{bulling2014tutorial}, which uses a sliding window-based activity detection framework.
Based on our analysis, the labeled target behavior episodes lasted 3 seconds on average.
We used a 4-second ($T=120$ frames for 30Hz videos) sliding window for modeling  activities in the group.
Since our video cameras capture personally identifiable details, respecting the privacy of individuals was a big priority.
Previous research \cite{yin2021see} has shown that gradients from a trained model can be used to reconstruct the training images.
Hence, for each analysis window, we first applied multi-person 2D pose estimation techniques ~\cite{geng2021bottom} and multi-person tracking techniques to collect individual activities from the group, but without using sensitive information, such as participants' faces~\cite{kuhn1955hungarian,Bewley2016_sort}.
Each person's activity trajectory is processed with a person-level activity feature extraction model, in which all individual activity features are aggregated to extract a video-level feature representation. 
We perform binary classification on the video-level feature representation to detect the presence of a target behavior in the analysis window.
Specifically, our model applies attention mechanisms~\cite{vaswani2017attention} to automatically identify the participant in the group, who might be most relevant to detecting episodes of a target behavior in the scene.
This is especially useful since target behaviors may be displayed by only a subset of children in the group.
Furthermore, we apply the body joint attention and temporal attention techniques to individual-level features for effectively modeling the most relevant body movements and time segments relating to target behaviors in the time window.
In the following sections, we discuss the details of our pipeline for multi-person detection and tracking, person-level representation, and video-level representation used in this work.

\subsubsection{Multi-person 2D Pose Estimation, Tracking, and Preprocessing:}
% \hyeok{This is the first part your method. Explain in detail.}
For each frame,  we used DEKR~\cite{geng2021bottom} for multi-person pose estimation, which is a state-of-the-art human pose estimation model that detects 17 body keypoints~\cite{lin2014microsoft} from nose to ankles associated with $N$ number of individuals in the frame $t$, $P^{1:N}_t\in\mathbb{R}^{N\times 17\times 2}$.
Then, for the 2D poses detected over the 4-sec window, $P^{1:N}_{1:T}\in\mathbb{R}^{T\times N\times 17\times 2}$, we apply a tracking algorithm based on Hungarian matching~\cite{kuhn1955hungarian,Bewley2016_sort} to acquire activities of individuals.
The Hungarian matching algorithm finds matches between the poses from two consecutive frames, $P^{1:N}_{t:t+1}\in\mathbb{R}^{2\times N\times 17\times 2}$, by generating a bipartite graph, represented as a matrix, $C$, where each node is the set of 2D poses from times $t$ and $t+1$, and the edge is the $L_2$-norm distance between those 2D poses, $C_{i,j}= \|P^i_{t} - P^j_{t+1} \|_2$.
The Hungarian method then finds the matchings (or edges) with the smallest sum of distances between the samples from $t$ and $t+1$, by optimizing $Y = \arg\min_{Y} \sum_i \sum_j C_{i,j} Y_{i,j}$, where $Y$ is a boolean matrix, showing the assignments within $C$. 
Due to ghost detections (false positives) in between frames, we found that very short ghost trajectories ($T<1$ sec) are detected.
Therefore, we only consider trajectories that are longer than 1 second, which are more likely to be actual subjects observed in the video.
In addition, we also hip-center and normalize the 2D pose coordinates for each detected person within a window. 

\subsubsection{Person-level Activity Representation}
% We used a convolutional neural network backbone to study the efficacy of attention mechanism in detecting clinically significant results. 

Our pipeline first learns each individual's activity representations over the 4-second windows from the pose sequence of each individual, $P^i_{1:T}\in\mathbb{R}^{T\times 17 \times 2}$.
The pipeline to learn person-level representations has three parts: 1) Body Joint Attention, 2) Temporal Feature Extraction, and 3) Temporal Attention and Pooling.

\textit{i) Body Joint Attention} model (\textit{J-Att}) processes 2D pose sequences of individuals and outputs attention weights that represent the relevance of joint movements, such as repetitive movements, hand clapping, or jumping, to target behaviors in ASD.
Specifically, we apply a two-layer temporal convolutional network (TCN) with two fully connected layers for processing 2D pose sequences.
The TCN has a filter size of $5\times 1$ with 64 feature maps, while the fully connected layers have 512 and 128 hidden units.
The joint attention output, $A_{joint}^{1:T}\in\mathbb{R}^{T\times 17\times 1}$, where $\sum_{i} A^t_{joint_i} = 1$ and $A_{joint}^{1:T}\geq 0$, is used downstream to weight temporal pose features before temporal pooling.

\textit{ii) Temporal Feature Extraction} model uses a four-layer TCN to extract temporal features, $X\in\mathbb{R}^{T\times 17 \times 256}$, over the pose sequences.
The temporal features are weighted with joint attention weights from \textit{J-Att} model, $\hat{X} = A_{joint}^{1:T}X$, which are utilized for learning joint-weighted temporal features.

\textit{iii) Temporal Attention and Pooling} model (\textit{T-Att}) aggregates the joint-weighted pose sequence features, $\hat{X}\in\mathbb{R}^{T\times 17 \times 256}$ over time based on the temporal importance weight.
\textit{T-Att} processes $\hat{X}$ using a two-layer TCN, which has a filter size of $5\times 1$ and 64 feature maps and one-layer fully connected layer with 128 hidden units to output temporal attention weights, $A_{time}\in\mathbb{R}^{T\times 1}$, where $\sum_t A^t_{time} = 1$ and $A_{time}\geq 0$.
This captures the most relevant frames within a 4-second window for identifying episodes of target behaviors in ASD.
Next, $\hat{X}$ is temporally aggregated over time using $A_{time}$, capturing the overall activity representation of each individual over the 4-second window, $\Tilde{X} = A_{time}\hat{X}$, where $\Tilde{X}\in\mathbb{R}^{17\times 256}$.
Model weights used in person-level representation learning are shared among all the participants in our dataset to learn a generic representation that captures behaviors within a classroom environment.

\subsubsection{Video-level Activity Representation}
We derive a video-level activity representation over the 4-second window containing $N$ individuals by aggregating their activity representations, $\Tilde{X}^{1:N}\in\mathbb{R}^{N\times 17\times 256}$, based on the relevance of the individual's features.
Although our goal is to detect the episodes of target behaviors in students with ASD, the participants in the video include not only children with ASD but also staff members and teachers.
Additionally, when episodes of a target behavior are observed, only a subset of children (or a single child) may be exhibiting those behaviors.
Therefore, it is important for our model to automatically identify the most relevant individual from the group with respect to the observed target behavior.
We adapt the attention model, namely the \textit{Person Attention} model (\textit{P-Att}), to learn the person-importance weight based on $\Tilde{X}^{1:N}$.
Specifically, we use a two-layer fully-connected model with 1024 and 256 hidden units to process $\Tilde{X}^{1:N}$ to derive $A_{person}\in\mathbb{R}^{N\times 1}$, where $\sum_i A^i_{person} = 1$ and $A_{person}\geq 0$.
Then, the video-level activity representation is the aggregation of $\Tilde{X}$ using $A_{person}$, which is $\dot{X} = A_{person}\Tilde{X}$, where $\dot{X}\in\mathbb{R}^{17\times 256}$.
Finally, the video-level activity representation, $\dot{X}$, is processed with a two-layer fully-connected model with 1024 and 256 hidden layers for a binary classification task to detect the presence of target behaviors in the analysis window.

In summary, the proposed model uses attention mechanisms at three levels, which are body joint movement, time window, and individuals.
These attention weights provide interpretability on the most relevant signal at each level for the given analysis window.
In our analysis, we will demonstrate specific activity types our model considers relevant for identifying target behaviors.

\subsection{Experiments}

\subsubsection{Model Training and Evaluation}

The proposed model is validated with a 5-fold cross-validation approach, and 50\%, 20\%, and 30\% of the dataset was used for the training, validation, and testing sets, respectively.
At each fold,  we avoided placing adjacent analysis windows in different folds to avoid pairwise similarity from biasing the cross-validation results~\cite{hammerla2015let}.
For the training model, we used a learning rate fixed at $1\times 10^{-3}$ with the Adam optimizer and a batch size of 16. 
Model training was stopped when no decrease in loss was observed from the validation set, with the same model being used for evaluating the test set.
Following the standard evaluation in human activity analysis for heavily imbalanced datasets, we used the F1-score, which is a harmonic mean of precision and recall, for model evaluation.
We also used the 95\% confidence interval of the F1-score for evaluating statistical significance in performance differences between the varying models described below.

\subsubsection{Baseline Models}

Our model has attention models at joint movement (\textit{J-Att}), time (\textit{T-Att}), and person (\textit{P-Att})-levels, and we will compare the proposed model with multiple baselines with and without attention models. 

\begin{enumerate}[label=\roman*)]
  \item \textit{TCN}: Standard temporal convolutional network without attention pooling models. This model does not have \textit{J-Att} and considers each joint movement equally important. It replaces \textit{T-Att} and \textit{P-Att} with simple global average pooling~\cite{lin2013network}, considering all timesteps and persons in the scene equally important.
Specifically, temporal features from TCN are mean-pooled across time, $\Tilde{X} = \frac{1}{T}\sum_t X_t \in\mathbb{R}^{17\times 256}$, to obtaion person-level representations. 
Also, the individual features, $\Tilde{X}^{1:N}$, detected in the scene are average pooled, $\dot{X} = \frac{1}{N}\sum_i \Tilde{X}^i$, where  $\Tilde{X}^i \in \mathbb{R}^{17\times 256}$.

  \item \textit{P-Att}: This model only applies \textit{P-Att} without \textit{T-Att} and \textit{J-Att}. Likewise, above, the temporal features are mean-pooled across time.
  
  \item \textit{PT-Att}: This model applies \textit{P-Att} and \textit{T-Att}, but not \textit{J-Att}.
\end{enumerate}

We compare the above models with varying complexities to our proposed model, \textit{PTJ-Att}.
We also compare the computing time for each model. 
The computing time for our model is evaluated by averaging the inference time over 50 runs on a set of 14 clips with varying numbers of subjects (4-7 individuals) in the scene.
Inference tasks were evaluated on a single NVIDIA Quadro RTX 6000 GPU with 24GBs of VRAM using a batch size of 16.
% \hyeok{Barun, fill in with CPU, GPU specs and number of runs for getting the average time. And, I believe the computing time includes 2D pose estimation, multiperson tracking, and inference. Is this correct?}.

\subsubsection{Detection and Prediction of Target Behaviors}
With the proposed method, we studied both detection and prediction of target behaviors continuously over the videos.

\begin{enumerate}[label=\roman*)]
    \item \textit{Detection task}: Considering future deployment scenarios for our model, we labeled the 4-second analysis window to contain a target behavior if the frame in the last timestep of the window is annotated as one of the target behavior categories. 
    This is to ensure our model can immediately detect target or other behaviors as soon as the model observes the onset or offset of an episode when processing video continuously using a sliding window manner.

    \item \textit{Prediction task}: We specifically studied if our model can predict the episodes of target behaviors 3 minutes into the future. 
    We trained and evaluated the model similar to the detection task.
    We process a 4-second %\barun{changing to 4-second} 
    window but with binary classification of behaviors observed 3 minutes after the last timestep of the analysis window. 
    The 3-minute prediction horizon is set in consultation with the staff at TCFD, considering that this is the desirable time horizon needed to intervene if a child exhibits high-risk target behaviors in the classroom.

    \item \textit{True Positive Rate per Behavior Category}: We also evaluated the True Positive Rate (TPR) of detecting a target behavior in each category discussed in \autoref{fig:problem-behaviors}.
    This is to study the suitability of the model for detecting different behaviors from video-based group activity analyses.
    % \barun{
    % }.
    % \sout{which behavior category is more suitable for a model to learn from video-based group activity analysis}.
    The proposed binary classification model infers whether the given analysis frame belongs to the target behavior episode or not.
    However, multiple children can exhibit more than one target behavior in a frame. In such a case, we considered each category of target behaviors to be detected (true positives) if the model infers that the analysis frame is present with the behavioral episode.
    % \barun{I am not sure I understand what this sentence is stating}.
\end{enumerate}

% \subsubsection{Effect of Personalization}
% \hyeok{Barun, fill the description for how you do in. 1. t-sne visualization of a video 2. pose visualization from clustering analysis}
% \barun{inconclusive}

\subsubsection{Visualizing Target Behaviors in ASD}\label{subsec:method-bh-phenotypes}

Our video analysis model is devised with attention models that can automatically detect the most relevant individual for episodes of target behaviors in classroom group activities.
We qualitatively evaluate the kinds of activities detected by the model for target behavior episodes with high confidence.
% \hyeok{Barun, fill in the details with your qualitative analysis}
% We devised an automated pipeline to extract representative behavior phenotypes for different behavior labels.
Specifically, we show the highlighted behavior types for the analysis windows belonging to \textit{repeated restrictive behaviors} and \textit{disruptive behavior}, as these were the largest amount of samples compared to other target behavior categories (\autoref{fig:problem-behaviors}).
We collected and processed all the windows in each category to extract person-level attention scores, $A_{person}$.
Then, we collected the person-level features with the highest attention score, $\Tilde{X}$.
The collected person-level feature representing the category was clustered with K-medoid to detect the most representative activity representations, which are at the cluster centers.
We assume that person-level activity representation that belongs to the cluster center of the largest cluster is the most representative sample for that category in our dataset.
Lastly, we visualized the frames and attended person corresponding to the representation.

    \section*{Results}

\subsection{Detection and Prediction of Target Behaviors}
% For the behavior detection task, the classifier was designed to output a binary label depending on whether a problem behavior was detected in the input window.
% We report the mean F1-score and 95\% confidence intervals after a 5-fold cross validation on our dataset, along with the inference time in seconds  in \autoref{tab:results-bce}

\begin{table}[t]
    \centering
    \caption{\small Target behavior detection and prediction results with different attention mechanisms.
    For the prediction task, we only tested with \textit{P-Att} model, as this showed the best performance in the detection task.
    (\textbf{Bold} text represents the score for the best performing model.)
    % The baseline model, without any attention heads, represents the performance of a convolutional network in this problem setting.
    % All results were derived after a five-fold cross validation 
    % \hyeok{This cross validation split does not give any useful information to understand this table. Remove it.}\textcolor{yellow}{(check percentages)}
    % \hyeok{\sout{Update the title. Binary detection does not mean anything. What are you detecting? Refer to my comments in the main text.
    % You don't need 4th decimals. They are just noise.
    % Remove the first column with row index. Does not add any information.
    % Add computational time and also discuss it in the main text.
    % }}
    % \resolved
    }    
    \begin{adjustbox}{width=\linewidth}
        \begin{tabular}{ccccccc}
            \multirow{2}{4em}{Models} & \multicolumn{3}{c}{Attention Heads} & \multicolumn{2}{c}{F1-score (\textit{95\% CI})} & Runtime(s)\\
            \cmidrule(r){2-6}
            & \begin{tabular}{@{}c} Multi-person \\ attention \end{tabular} 
            & \begin{tabular}{@{}c} Temporal \\ attention \end{tabular} & \begin{tabular}{@{}c} Joint \\ attention \end{tabular} 
            & \begin{tabular}{@{}c} Cam 1 \\ (Top-down) \end{tabular} 
            & \begin{tabular}{@{}c} Cam 2 \\ (Side) \end{tabular} 
             \\
            \hline
            \textbf{Detection} & & & & & &\\
            % \textbf{Classification}
            
            \textit{TCN} & - & - & - & 0.74 $\pm$ 0.01 & 0.52 $\pm$ 0.07 & 50.33\\
            \textit{P-Att} & \checkmark & - & - & \textbf{0.77 $\pm$ 0.01} & \textbf{0.52 $\pm$ 0.07} & 60.24 \\
            \textit{PT-Att} & \checkmark & \checkmark & - & 0.73 $\pm$ 0.07 & 0.48 $\pm$ 0.01 & 113.06\\
            \textit{PTJ-Att} & \checkmark & \checkmark & \checkmark & 0.73 $\pm$ 0.01 & 0.51 $\pm$ 0.06  & 184.53 \vspace{4pt} \\
            \hline 
            \textbf{Prediction} & & & & & & \\
            \textit{P-Att} & \checkmark & - & - & 0.53 $\pm$ 0.07 & 0.51 $\pm$ 0.06 & 
        \end{tabular}
    \end{adjustbox}
    \label{tab:results-bce}
\end{table}
\begin{figure*}[t!]
    \centering
    \includegraphics[width=\textwidth]{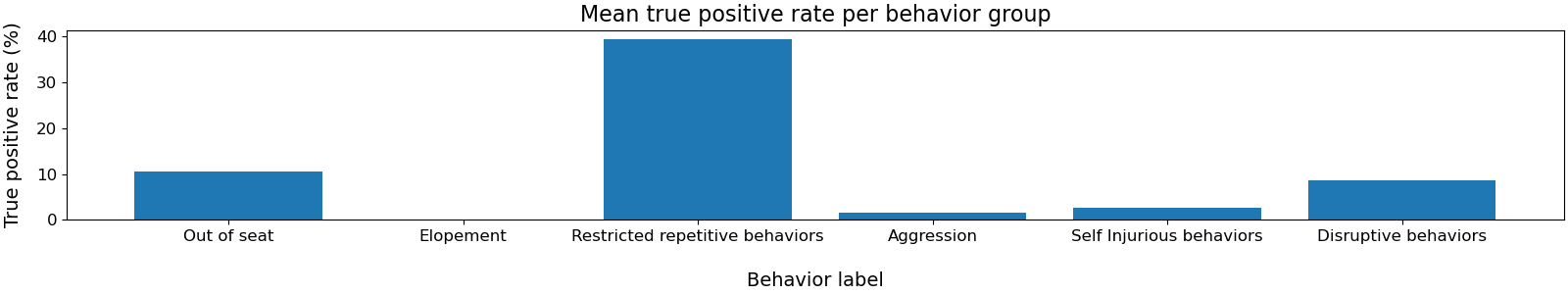}
    \vspace{-2em}
    \caption{Mean true positive rate (TPR) for different behaviors. TPR for each behavior is correlated with the availability of the sample belonging to each category from \autoref{fig:problem-behaviors}}
    \label{fig:tpr}
\end{figure*}

Overall, our model identified target behaviors during group activities with a 77\% F1 score when using top-down videos, despite having multiple bystanders, including TCFD staff or teachers, in the scene.
\autoref{tab:results-bce} shows the F1-score for the models with different attention mechanisms for both top-down (Cam 1) and side (Cam 2) views from \autoref{fig:camera-views}.
For detection tasks (1-4 rows), the model with person attention (\textit{P-Att}, 2nd row) for top-down view (Cam 1) demonstrated best performance (0.77 F1-score) showing absolute improvements of 3\% compared to the model without any attention (\textit{TCN}, 1st row), while maintaining a similar inference time with only a 10 seconds difference.
Due to the higher complexity of models, \textit{PT-Att} (3rd row) and \textit{PTJ-Att} (4th row) models required $2\times$ and $3\times$ the inference time compared to \textit{TCN} and \textit{P-Att}.
% From the results on \autoref{tab:results-bce}, it is apparent that the baselines and \textit{P-Att} models run within a 1 minute window, while the \textit{PT-Att} and \textit{PTJ-Att} models take almost twice longer for the same task. 
Prediction tasks (5th row) were only evaluated with the best-performing model from the behavior detection task, i.e., \textit{P-Att}. 
Predicting target behaviors within a 3-minute horizon is demonstrated to be very challenging, showing a 0.53 F1-score for \textit{P-Att} model with a top-down view, which is only slightly higher than random chance for a binary classification problem. 
\autoref{fig:tpr} shows the true positive rate for each behavior category.
This was evaluated using the best performing model, i.e., \textit{P-Att} on Camera 1 videos.
The \textit{restricted repetitive behaviors} category had the highest true positive rate at 39.48\% and \textit{elopement} had the lowest true positive rate at 0.05\%.

\subsection{Identifying Target Behaviors in ASD}
\begin{figure}[t]
    \centering
    \begin{subfigure}[b]{\textwidth}
        \centering
        \includegraphics[width=0.9\linewidth]{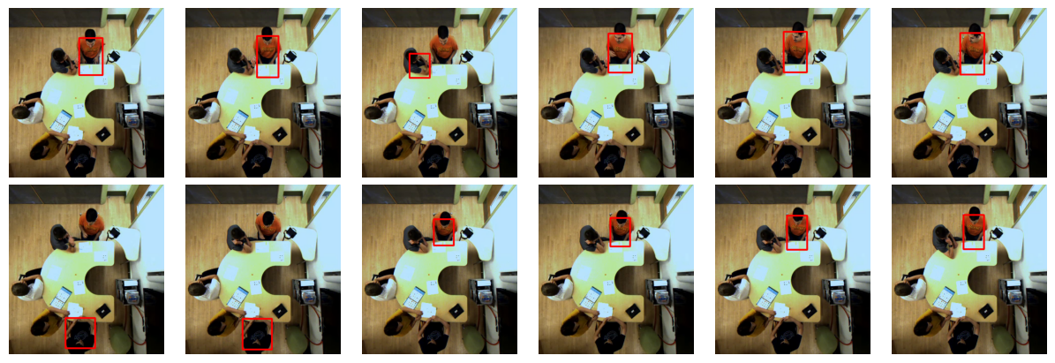}
        \caption{Behavior phenotypes for \textit{repeated restrictive behaviors}, where a subject is jumping up and down.}
        \label{fig:behavior-phenotype-rep}
    \end{subfigure}
    
    \vspace{0.2em} % Add vertical space between the images
    
    \begin{subfigure}[b]{\textwidth}
        \centering
        \includegraphics[width=0.9\linewidth]{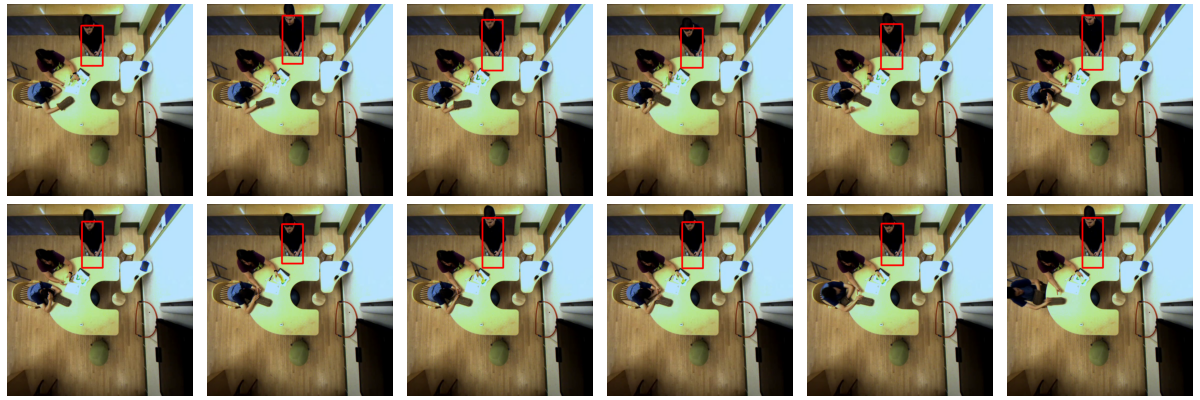}
        \caption{Behavior phenotypes for \textit{disruptive behaviors}, where a subject is rotating head left and right.}
        \label{fig:behavior-phenotype-disruption}
    \end{subfigure}
    
    \caption{Example behavior phenotypes detected by the P-Attn model.
    12 frames from a 4-second window for a most representative sample is displayed with a bounding box highlighting the subject detected by the model to show target behavior.
    \autoref{fig:behavior-phenotype-rep} highlights a subject jumping up and down labeled as \textit{repeated restrictive behavior and \autoref{fig:behavior-phenotype-disruption}} highlights a subject rapidly rotating the head left and right labeled as \textit{disruptive behavior}.
    %--
    % Representative behavior phenotypes with bounding boxes around the individual most responsible for the detected behavior signal in different problem behaviors observed among children. 
    % % We picked 12 equally spaced frames from the most representative 4s window of the behavior. 
    % \autoref{fig:behavior-phenotype-best} represents phenotypes from the behavior label with the highest true positive rate of detection and \autoref{fig:behavior-phenotype-worst} is the same for the behavior label with the lowest true positive rate of detection.
    % \barun{Panels need to be anonymized}
    }
    \label{fig:bh-phenotypes-r-d}
\end{figure}

% \hyeok{TK}

\autoref{fig:bh-phenotypes-r-d} shows the detected target behavior types by the \textit{P-Att} model. 
12 frames from 4-second analysis windows are shown for \textit{repeated restrictive behavior} and \textit{disruptive behavior} examples. 
The \textcolor{red}{red} bounding box shows the attended person within a 4-second analysis window.
The most representative analysis window for \textit{repeated restrictive behavior} and \textit{disruptive behavior} were jumping up and down in a chair and rotating head left and right, respectively.

% \hyeok{Barun, add a figure and brief intro of results. We will discuss this further in the Discussion part}
% As described in the Experiments section, we developed an automated pipeline which computes the most representative window $W_B$ for a behavior label $B$.
% We then generated bounding boxes for the person with the highest attention score in that window.
% However, it is important to note that there were very few labeled windows for certain atypical behavior labels such as \textit{Aggression} and \textit{Elopement} due to label sparsity and imbalance in our dataset.
% As a result, behavior phenotypes detected for these labels are expected to be highly biased.

    \section*{Discussion} 
\label{sec:discussion}

% \hyeok{The discussion has to first start with overall result of the experiment. What was the best performance? If you divide those binary detection into each activity class, which activity the model missed or accurately detected? Then, you will need to go to how the performance was different according to the camera view point. Why? because camera view point is a bigger factor that makes performance difference than different model variants. Afterwards discuss the model variant performances using Cam 1. Now that all quantitative discussion is done move to the qualitative analysis. The MAJOR qualitative analysis is the activity phenotypes and visualization of movements based on clustering analysis, which we discussed multiple times. Have sufficient portion for this discussion you can have one good figure and discussion can be up to 1.5 page. Longer is okay. We can later pick, which one to cut down. So far you have finished all discussions relating to detection task, and now briefly discuss the prediction result. why the prediction was challenging? What do you think is next step to overcome the challenges? Next, discuss the limitations of this work including dataset and that our model does not necessarily guarantee identifying which child is showing problem behavior and also limited in identifying the activity class. This naturally leads to discussing your future work. You will need to fully revise it. For now, I will put my comments based on this flow.}

\subsection{Detection and Prediction of Target Behaviors}

\subsubsection{Camera View}

\autoref{tab:results-bce} shows that the camera viewpoint has a significant impact on analyzing group activities in the classroom and, for \textit{P-Att},  side view (Cam 2) resulted in a 25\% absolute decrease in the F1-score.
From our observations, the side view had frequent occlusions of children's activities in the classroom, especially when children showed target behaviors and moved out of their seats.
On the other hand, a top-down view (Cam 1) could capture detailed movements of the children when they exhibited target behaviors. 
Notably, it clearly captured children’s upper limb movements as well as instances when they moved out of their seats.
We consider that the improved visibility of children's movements significantly helped us to learn representations of target behaviors and identify the children exhibiting the behavior.

\subsubsection{Attention Models}

We originally hypothesized that the more flexible model would be better at learning behavior representations of target behaviors.
However, the \textit{PT-Att} and \textit{PTJ-Att} models performed similarly to the \textit{TCN} model without any attention modules, with only the \textit{P-Att} outperforming it.
We consider that more complex models (\textit{PT-Att} and \textit{PTJ-Att}) were overfitting due to the small dataset size. 
In this work, we only had 22 hours of data, and less than 20\% of the frames ($\sim 4$ hours) showed target behaviors, resulting in a highly imbalanced dataset with high heterogeneity, having the six different categories of atypical behaviors illustrated in \autoref{fig:problem-behaviors}.
% We consider the high flexibility of the \textit{PT-Att} and \textit{PTJ-Att} models led to the overfitting, resulting in a lower performance compared to the \textit{P-Att} model.
However, the statistically significant improvements of the \textit{P-Att} model over the \textit{TCN} model highlight the importance of the \textit{P-Att} model in automatically identifying individuals exhibiting atypical behavior, even in scenes involving other people, such as staff, teachers, or children displaying typical behaviors.
% Yet, the statistically significant improvements of \textit{P-Att} model over \textit{TCN} model indicates the importance of the \textit{P-Att} model that can automatically identify the person relevant to the target behavior when the scene includes other individuals, such as staff, teachers, or children with other behaviors.
Additionally, behavior prediction is demonstrated to be a very challenging task, as shown by the performance of the \textit{P-Att} model.
Our goal of predicting activity 3 minutes into the future is very ambitious, given that previous work in human activity or kinematics prediction has demonstrated that the prediction of human behavior, in general, is very challenging beyond 3 to 5 seconds~\cite{guan2020generative,qi2017predicting,kong2022human,piergiovanni2020adversarial}
% , which makes our goal for predicting activity 3 minutes into the future similar to random guess. 
% \barun{(replace with): highly ambitious to begin with}.
In this study, we only considered 3 minutes for the prediction task, as it was determined that this was a substantial amount of time for our staff to intervene in case of a severe behavior (e.g., self-injurious behavior). 
Shorter time horizons will be considered in future work, given that human reaction times rely on several factors and are contextually dependent. Further validation is required to understand what are the minimum and optimum prediction widows to allow for effective behavioral intervention, as well as the time horizon at which the model retains the highest amount of accuracy while still remaining clinically useful. 
% \barun{Does this sentence diminish our work? Would the reviewers not question why we chose to include such a result in this case, if it is already known to be not much better than a random guess?}

\subsubsection{Detection per Behavior Category}

\autoref{fig:tpr} shows that our trained model, \textit{P-Att}, is highly biased in detecting each behavior category due to the bias in our dataset. 
Compared with \autoref{fig:problem-behaviors}, the TPR for each category closely follows the ratio of available labeled frames per behavior category.
The majority of samples in our dataset being \textit{restricted repetitive behaviors} led to the highest performance for the corresponding category.
The \textit{elopement beahvior} had the lowest TPR due to the least amount of samples in our dataset.
The highest TPR (\textit{restricted repetitive behaviors}) was still below 40\%.
We consider this was due to the highly imbalanced ratio of target vs. other behavior in our dataset, where 80\% of our dataset consisted of other behaviors.
Yet, this result shows potential to identify fine-grained categories of target behaviors from real-world classroom activities in children with ASD once more labeled datasets are available, which is a part of our future work.

\subsection{Detecting Target Behaviors in ASD}

% \hyeok{Barun, revisit this after the result. Add best and worst category of atypical behavior to detect and predict based on binary classification.}

\begin{figure}[t]
    \centering
    \includegraphics[width=0.9\linewidth]{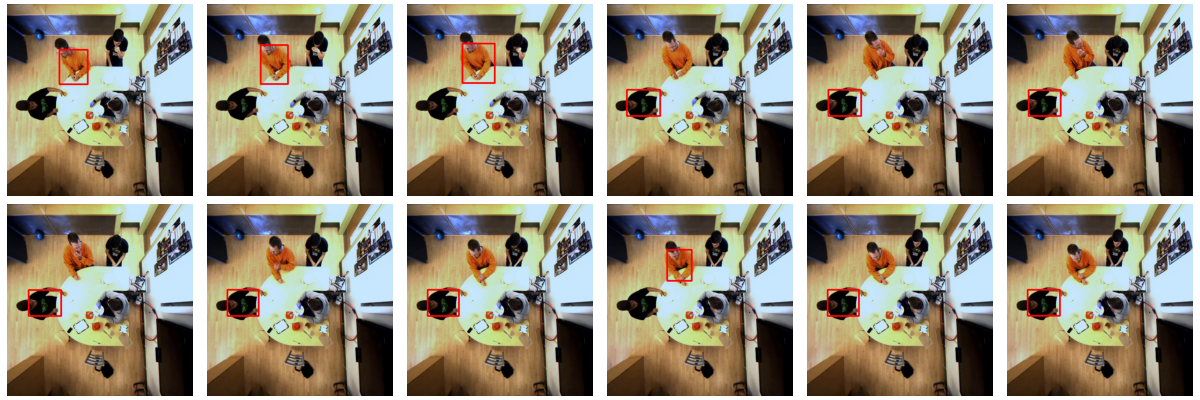}
    \caption{Failure case for detecting behavior phenotypes.
    Since the model only considers activity context and person observed within the window, it wrongly highlights the one student and one staff to predict \textit{elopement}, failing to capture the empty chair.}
    \label{fig:bh-phenotypes-ous}
\end{figure}

The proposed \textit{P-Attn} model could accurately and automatically detect the child displaying interfering or high-risk behaviors. 
For \textit{restricted repetitive behavior} and \textit{disruptive behaviors}, our model's attention results highly aligned with annotations from research assistants.
However, it also comes with a limitation.
Our model is designed to attend to the people detected by our multi-person 2D pose estimation model. 
As a result, this model fails to find relevant behavior types when the child exhibiting a target behavior is not present in the scene, such as in \textit{elopement} or \textit{out-of-seat} behaviors.
\autoref{fig:bh-phenotypes-ous} shows the individuals attended by our model when detecting behavior types for \textit{elopement}.
The scene clearly shows the empty seat due to \textit{elopement}. However, our model attends to two other children who are showing other non-target behaviors in the analysis window.
This shows that the model requires to be extended to consider the absence of children in the analysis window.

% Since we have multiple atypical behavior labels, in this section we highlight the behavior phenotypes detected for the best and worst performing categories to present a holistic view of our model.
% Accordingly, we analyzed the true positive rate of atypical behavior detection, as shown in \autoref{fig:tpr}.

% \autoref{fig:bh-phenotypes} shows behavior phenotypes detected by the \textit{P-Att} model for behaviors \textit{Restricted repetitive behaviors} and \textit{Elopement}, which were the behaviors with the highest and lowest true positive rate respectively.
% However, the \textit{Elopement} label only had 52 analysis windows in the entire dataset, making any conclusions drawn from it highly prone to bias.
% \barun{Completed}

\subsection{Future Work}

Our results in \autoref{tab:results-bce} show that the detection of interfering or high-risk target behaviors in group activities is possible.
However, multiple limitations remain that need to be addressed to validate the proposed framework's generalizability in the future.
Firstly, we have only 22 hours of labeled data, which is quite small.
We have an unlabeled dataset with 4 years of video data collected from daily classroom activities, which can be transformed into a rich resource to validate our system for longitudinal and large-scale evaluation.
Recent works in active learning frameworks~\cite{ren2021survey} have demonstrated that human and AI interaction systems can significantly decrease the cost of manual data annotation through continuous calibration (or training) of machine learning models. This is done by providing human inputs exclusively for the few samples that are challenging to classify.
We aim to utilize the trained model from this work to develop an active learning framework in a future project, which will enable efficient interactions with our clinical research team for labeling the entire dataset collected so far.
The large-scale labeled dataset is expected to unlock the potential of our proposed model, \textit{PTJ-Att}, to effectively learn behavior features in ASD.
The trained model with expert calibration will also be deployed across multiple classrooms to continuously monitor target behaviors in children of varying age ranges and validate the proposed model across different settings and groups of individuals.

Furthermore, the current model only identifies the presence of target behavior episodes, although there exist multiple categories of interfering and high-risk behaviors displayed by children with ASD from \autoref{fig:problem-behaviors}. 
As shown in \autoref{fig:subject-behaviors}, different children are likely to exhibit different frequencies of target behaviors.
Therefore, it is important to understand fine-grained target behaviors and their changes over time for children with ASD.
In our future work, we will expand our model to classify between 6 types of interfering and high-risk target behaviors. 
We also plan to integrate facial recognition models for side views with the top-down view model to identify specific kinds of target behavior exhibited by children.
We will also expand our video analysis model to a multi-modal system using ambient audio data captured by our video system and consider the activity context beyond 4 seconds to quantify both movement and speech behaviors in children with ASD.

    \section*{Conclusions}

Continuous monitoring of target behaviors is important for determining treatment effects and for tracking the overall progress of children with ASD. 
We developed a video-based explainable artificial intelligence technique that analyzes group activities in real-world classroom environments.
Our model could detect episodes of interfering or high-risk target behaviors with a 77\% F1 score using cameras in a top-down view and the proposed person attention mechanism, \textit{P-Attn}, could automatically identify a specific child exhibiting target behaviors.
These results show the feasibility of deploying camera-based analysis systems in classrooms, which can reduce costs and alleviate the staff burden of manually collecting behavior data through conventional practices. Educational settings and classroom ratios typically do not allow for separate observers to collect behavior data, especially continuous data. Time sampling methods may be more practical, but can over or underestimate behaviors depending on whether partial or whole interval recording is used~\cite{cooper2000adaptive}. Staff members are often expected to keep track of metrics like frequency and duration of behavioral episodes, while also having to manage behaviors and ensuring safety. This often results in a delay between the occurrence of the behavior and recording data, which decreases accuracy.  Automated detection of behavioral events would not only free up staff members to focus on instruction and support, but could also result in more accurate data collection.  

This is an important step for large-scale and longitudinal studies for passively and objectively monitoring behaviors in children with ASD participating in diverse classroom activities.
In our future work, we will further develop a multi-view subject identification model using facial recognition techniques and further extend our framework to model audio-visual behaviors for detailed detection of the target behaviors in \autoref{fig:problem-behaviors}.
This model will be integrated with low-cost edge computing devices, such as a Raspberry Pi or an Nvidia Jetson, for a cost-effective deployment of our model across multiple classrooms in the TCFD educational program~\cite{kwon2023feasibility}.

}

\ifthenelse{\boolean{blinded}}
{
}{
    \begin{acks}
    This work was supported by the Center for Discovery. Ali Rad, Gari Clifford, and Hyeokhyen Kwon were partially supported by the James M. Cox Foundation and Cox Enterprises, Inc., in support of Emory’s Brain Health Center and Georgia Institute of Technology. Gari Clifford is partially supported by the National Center for Advancing Translational Sciences of the National Institutes of Health under Award Number UL1TR002378. Hyeokhyen Kwon is partially supported by the National Institute on Deafness and Other Communication Disorders under Award Number 1R21DC021029-01A1.
    \end{acks}
}

\ifthenelse{\boolean{coverpage}}
{
}{
    % Reference
    \bibliographystyle{mslapa}
    \bibliography{reference}
}

\end{document}